%% file: iclr2022_conference.tex
\documentclass{article} 
\usepackage{iclr2022_conference,times}

\input{math_commands.tex}

\usepackage{hyperref}
\usepackage{url}
\usepackage{graphicx}
\usepackage{appendix}

\title{Evaluating the Adversarial Robustness for Fourier Neural Operators}

\usepackage{color}

\author{Abolaji D. Adesoji \thanks{ The author performed this research during his graduate studies at RPI but at the time of this publication, works for IBM} \\
Department of Mechanical, Aerospace, \\ and Nuclear Engineering,\\
Rensselaer Polytechnic Institute (RPI)\\
Troy, NY 12180, USA \\
\texttt{abolaji.adesoji1@ibm.com} \\
\And
Pin-Yu Chen \\
IBM Research \\
Yorktown, NY 10598, USA \\
\texttt{pin-yu.chen@ibm.com} 
}

\iclrfinalcopy 
\begin{document}

\maketitle

\begin{abstract}
In recent years, Machine-Learning (ML)-driven approaches have been widely used in scientific discovery domains. Among them, the Fourier Neural Operator (FNO) \citep{zongyi} was the first to simulate turbulent flow with zero-shot super-resolution and superior accuracy, which significantly improves the speed when compared to traditional partial differential equation (PDE) solvers. To inspect the trustworthiness, we provide the first study on the adversarial robustness of scientific discovery models by generating adversarial examples for FNO, based on norm-bounded data input perturbations. Evaluated on the mean squared error between the FNO model's output and the PDE solver's output, our results show that the model's robustness degrades rapidly with increasing perturbation levels, particularly in non-simplistic cases like the 2D Darcy and the Navier cases. Our research provides a sensitivity analysis tool and evaluation principles for assessing the adversarial robustness of ML-based scientific discovery models.
\end{abstract}

\section{Introduction}\label{sec:intro}
The recent data explosion and ML compute advancements has sparked research into ML's impact on scientific discovery. In lieu of this, researchers have built new ML models to learn and predict complex sciences. An example is seen in a class of ML models termed Physics-Informed Neural Networks (PINN) that parameterize the unknown solution $u$ (to avoid ill-conditioned numerical differentiations) and the nonlinear function $\mathcal N$ (to distill the nonlinearity to a spatiotemporal dataset), each with a Deep Neural Network (DNN) and is mesh invariant \citep{deepH}. In some light, it replaces the local basis functions in standard Finite Element Method with the neural network (NN) function space. Its drawbacks include its underlying PDE knowledge-need to aid its loss term formulation and its ability to only model a single instance of the system. Architecturally, training is done on the mean squared error (MSE) minimization for both networks.\\

Another class of models is the Neural operator which solve most of the drawbacks associated with prior models. They are mesh-free, infinite-dimensional operators that produce a set of network parameters that works for many discretizations but with a huge integral operator evaluation cost. \citet{zongyi} solved this by taking this operation to the Fourier space. This study explores the adversarial robustness of these Fourier Neural Operator (FNO) models, that is, the worst-case discrepancy between the FNO model's prediction and the ground-truth given by the corresponding PDE solver against norm-bounded data input perturbations.
To our best knowledge, the study of adversarial robustness for FNO has not been explored in current literature.\\

In this paper, we generate adversarial examples of the inputs to the FNO model for three cases studied in \citep{zongyi}: 1D Burgers, 2D Darcy, and 2D+1 Navier stokes equations. When evaluating the MSE v.s. the perturbation level $\epsilon$  on the generated adversarial examples, we find that the inconsistency between FNO models and PDE solvers diverge at a drastic rate as $\epsilon$ increases, suggesting that current FNO models may be overly sensitive to adversarial input perturbations.


\section{Fourier Neural Operator and Use Cases}\label{sec:FNO}

\citet{zongyi} introduces an efficient scientific discovery ML model that parameterizes the integral kernel in Fourier space. The kernel integral operator starts out as a linear combination, then becomes a convolution in the Fourier space. This mesh-invariant and physics-independent model also simulates turbulence with zero-shot super-resolution as it learns the resolution invariant solution operator, and is about 3 orders of magnitude faster at inference time than all other models considered.\\

For training, we have the data \{$a_j, u_j\}^N_{j=1}$, where $a_j \sim \mu$ is an i.i.d. sampled sequence of coefficients from $\mu$ and $u_j = G ^\dagger(a_j) $ is potentially noisy. The goal is to seek $G ^\dagger$ as the solution operator.\\
\begin{align}
&G_\theta: \mathcal{A} \rightarrow \mathcal{U}, \ \theta \in \Theta \\
v_{t+1}(x) &:= \sigma(Wv_t(x) + (\mathcal{K}(a;\phi)v_t)(x)) \label{eq:iter} 
\end{align}

Where $G_\theta$ is the solution operator in $\Theta$, the finite-dimensional parametric space. We seek to minimize the cost functional defined on it. The input and output are functions of the euclidean space, $x$ and time, $t$. Eq.~\ref{eq:iter} is the iterative update, and note that $v$ is the NN representation of $u$, $W$ is the weight tensor and the update is a composition of the non-local integral operator $\mathcal{K}$ which maps to linear bounded operators on $\mathcal{U}(D;\mathbb{R}^{d_u})$. $D$ is the domain ($D \in \mathbb{R}^d$). The Kernel Integral Operator becomes a convolution operator in the Fourier space:\\
 \begin{align}
 \Big (\mathcal K (&a;\phi)v_t\Big) (x) =  \mathcal{F}^{-1} \Big( \mathcal{F}  (\kappa_\phi) \cdot \mathcal{F} (v_t)\Big)(x)   \hspace{20px}\forall \ x \in D
\label{eq:kio} \\
&\Big (\mathcal K (\phi)v_t\Big) (x) =  \mathcal{F}^{-1} \Big( R _\phi \cdot \mathcal{F} (v_t)\Big)(x)   \hspace{20px}\forall \ x \in D \label{eq:fio} 
\end{align}

Where $\kappa_\phi$ is the linear operator, $\mathcal{F}$ is a Fourier transform and $R_\phi$ is the weight tensor to be learnt. We defer the details to \citep{zongyi}. Next, we introduce three PDE uses cases as studied by the FNO model \citep{zongyi}. The solutions generated from the PDE solvers will serve as the ground-truth for our robustness evaluation. 

\paragraph{1D Burgers case}\label{subsec:burgers}
This equation is for the viscous fluid flow with initial boundary condition (b.c.) $u_0 \in L^2_{per}((0,1);\mathbb R)$, where $L^2_{per}((0,1)$ is the 1D $L^2$ real space and $\partial$ is the differential operator.\\
\begin{align} 
\partial_t u (x,t) + \dfrac{\partial_x u^2(x,t)}{2} &= \nu \partial_{xx}u(x,t) \hspace{10px} x \in (0,1),~t \in (0,1] \\ 
&u(x,0) = u_0(x)
\end{align}

\paragraph{2D Darcy Flow} \label{subsec:darcy}
This is a steady state flow through a porous media unit box and Dirichilet b.c. is enforced.
\begin{align} 
- \nabla \cdot (a(x) \nabla & u (x)) = f(x) \hspace{30px} x \in (0,1)^2 \\ 
u(x) &= 0 \hspace{30px} x \in \partial (0,1)^2
\end{align}
Where $a$ is the diffusion coefficient, $u$ is the solution and $f$ is the forcing term. We seek to learn the mapping from the coefficients to the solution -- the PDE is linear but the operator $G^\dagger$ is not.

\paragraph{2D + 1 Navier stokes}\label{subsec:navier}
Given a 2D viscous incompressible flow with $u \in C([0,T]; \  H^r_{per}((0,1)^2,\mathbb R^2)$ as the velocity field for $r>0$ and $w=\nabla \times u$ as the vorticity, the model learns the vorticity field $w$ but at different time scales ($T< 10 \ $ vs $ \ T>10$). Where $H^r_{per}$ is the Hilbert space and $\nabla$ is the gradient operator.\\
\begin{align}
\partial_t w + u \cdot \nabla w = \nu \Delta w  + f; \ \forall \ u&(x,t) w(x,t) f(x) | x \in (0,1)^2,~t \in (0,T]\\
\nabla \cdot u = 0 \hspace{20px}&  x \in (0,1)^2,~t \in (0,T]\\
w(x,0) = w&_0(x)\hspace{20px}x \in (0,1)^2
\end{align}

\section{Adversarial Robustness Evaluation on FNO}\label{sec:adverse}
Adversarial attacks aim at finding failure modes of a given ML model \citep{biggio2018wild,goodfellow2014explaining,carlini2017towards,chen2022holistic}. We extend input perturbation (originally flourished within image classification models) to ML-based scientific discovery models, especially the FNO model. This research is poised to initiate adversarial robustness considerations into models deployed in the scientific discovery domains. Our study assumes the attacker has gradient (white-box) access and utilizes the Projected Gradient descent (PGD) attack, as introduced in \citep{madry2017towards}, for generating both $\ell_\infty$-norm bounded perturbations to data inputs.

\subsection{Proposed Method}\label{subsec:grid}

The traditional PDE solver only works with the nonsampled input field $\mathbf a_f$. Since the training of FNO is performed on the input field subsampled at a rate $s$, the attackers' objective is to attack a non-subsampled field that is almost identical to $\mathbf a_f$. We performed grid searches for the closest approximating surrogate fields from a larger collection of inputs. The pseudocode below is only used in the Burgers and Darcy case, due to their already subsampled training data, while the Navier case uses the traditional PDE \ solver directly. Note that $n$ is training set size, $N$ is the number of random Gaussian Random Fields (GRF) generated, $f$ is the full grid size, $d$ is the number of dimensions and $\delta$ is the perturbation.

\begin{enumerate}
\item  Perturb each of the already subsampled training input to give tensor $\mathbf a_s$ of size $s^{d}$
\item  Generate $N$ GRF, $\mathbf a_f$ of size  $f^d$  , where $n \ll N$
\item Stack these $N$ fields together to give the larger tensor $\mathcal{A}_f$ of size $N \times f^{d}$
    \item Subsample $\mathcal{A}_f$ at the rate $s$ to give the tensor $\mathcal{A}_s$
    \item Do a grid search through $\mathcal{A}_s$, for the $L_2$ distance minimizing field $\mathbf b_j$ closest to $\mathbf a_s$
    \item Complete the search for all $n$ perturbed input fields $\mathbf a_s$
    \item Stack the resulting surrogate fields $\{ \mathbf b_j \}$ to give the surrogate tensor $\mathcal B$ (approximate ground-truth) of size  $n \times s^d$
    \item Retrain the FNO model using $\mathcal B$ as our training tensor
    \item Save the full and subsampled output fields from the solver, for MSE calculation
\end{enumerate}

Given a data input $a$, the attacker's objective aims to find a perturbation $\delta$ with an $\ell_\infty$-norm bounded constraint $\|\delta \|_\infty \leq \epsilon$ to maximize the discrepancy between the FNO model's output $G_\theta(a+\delta)$ and the ground-truth (or the closest surrogate) from the PDE solver denoted by $g(a+\delta)$. The attack objective function is to maximize their mean squared error (MSE) defined as $\textsf{loss}:=\|G_\theta(a+\delta)-g(a+\delta)\|_2^2$.
We use the $\ell_{\infty}$ projected gradient descent (PGD) attack \citep{madry2017towards} to solve for $\delta$, which takes $K$ steps of gradient ascents followed by $\epsilon$ clipping using gradient sign values:
\begin{align}
    \delta^{(k+1)} := \textsf{Clip}_{[-\epsilon,\epsilon]} \left( \delta^{(k)} + \alpha \cdot \text{sign}(\nabla_{\delta^{(k)}} \textsf{loss} )\right)
\end{align}

\section{Performance Evaluation}\label{sec:results}

In this section, we show the FNO model's performance in all three aforementioned cases when adversarially attacked with different perturbation level $\epsilon \in (0.01,\ 5)$. The data input range of FNO models is unbounded. Also, the MSE computations are done with  $100$ randomly selected data samples from the test set for all cases, and the average MSE is used as the reported performance metric. Other setup parameters are listed in Appendix Section~\ref{subsec:params}.

\subsection{1D Burgers Equation Results}\label{subsec:burgers_result} 
\begin{figure}[ht]
    \begin{minipage}[c]{0.49\linewidth}
    \centering
    \includegraphics[  width=\linewidth,
    ]{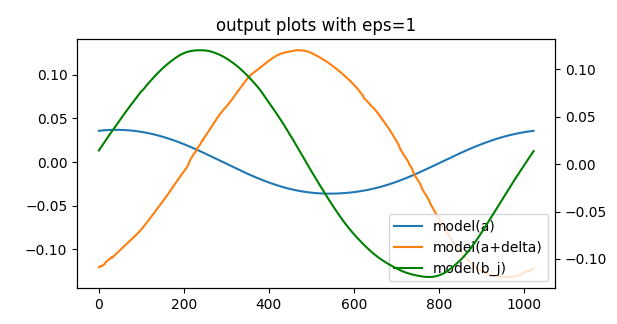}
    \caption{Burgers output field. \textbf{x-axis}: Grid location, \textbf{y-axis}: Output value} 
    \label{fig:grf_comp}
    \end{minipage} 
    \begin{minipage}[c]{0.49\linewidth}
    \centering
    \includegraphics[  width=\linewidth,
    ]{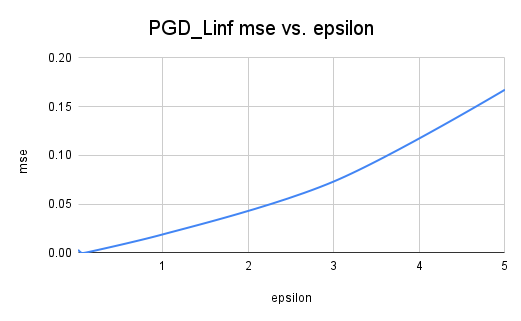}
    \caption{Burgers $\ell_\infty$-PGD MSE vs. $\epsilon$} \label{fig:pgd_linf_mse_burger}
    \end{minipage} 
\end{figure}

In line with the procedure outlined in Section~\ref{subsec:grid}, we trained an FNO model on perturbed 1D Burgers data using our proposed robustness evaluation procedure. Fig.~\ref{fig:grf_comp} shows the outputs of the FNO model and the solver on one instance of the original input data $a_j$ (abbreviated as $a$ in the plot), the perturbed data $a_j+\delta$ and the closest surrogate field $b_j$. The $b_j$ fields were chosen from 5,000 GRF realizations. Fig.~\ref{fig:pgd_linf_mse_burger} shows the $\ell_\infty$-PGD MSE variation with $\epsilon$, and it can be observed that the MSE rises at roughly 80\% faster than $\epsilon$, suggesting that the sensitivity of FNO intensifies rapidly with increased adversarial perturbation levels. For qualitative analysis and visualization, Fig.~\ref{fig:burgers_input} and Fig.~\ref{fig:burgers_output} in the Appendix show Burgers input and output fields before and after the attack. 

\begin{figure}[ht]
    \begin{minipage}[c]{0.49\linewidth}
    \centering
    \includegraphics[  width=\linewidth,
    ]{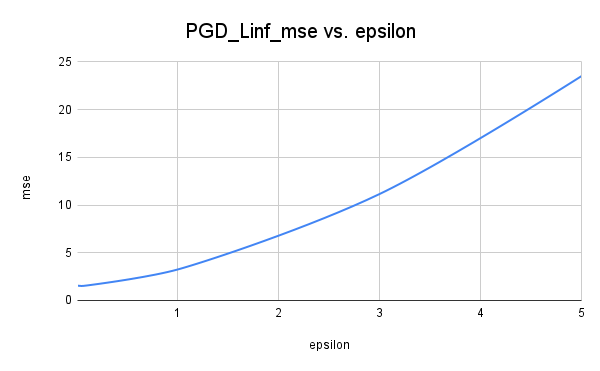}
    \caption{Darcy $\ell_\infty$-PGD  MSE vs. $\epsilon$}
    \label{fig:pgd_linf_mse_darcy}
    \end{minipage} 
    \begin{minipage}[c]{0.49\linewidth}
    \centering
    \includegraphics[  width=\linewidth,
    ]{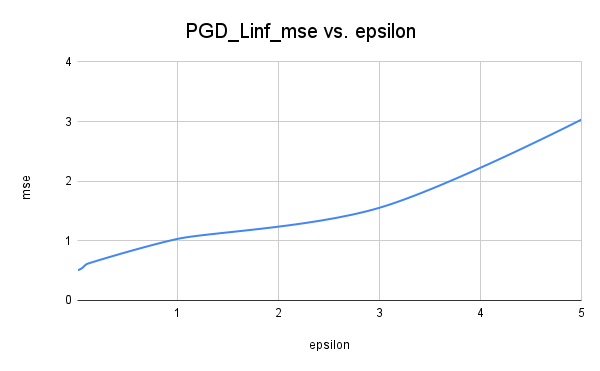}
    \caption{Navier $\ell_\infty$-PGD MSE vs $\epsilon$}
    \label{fig:pgd_linf_mse_navier}
    \end{minipage}     
\end{figure}

\subsection{2D Darcy Equation Results}\label{subsec:darcy_result}
Similar to~Section \ref{subsec:burgers_result}, we used a 2D surrogate subsampled grid search, to meet the solver's need for GRF of an acceptable spline format. Also, $\{b_j\}$ fields were selected from 10,000 realizations in an $L_2$-distance minimizing scheme, though it may not be enough to accurately match (visually) the perturbed fields. Fig.~\ref{fig:pgd_linf_mse_darcy} shows the MSE's seemingly quadratic relationship with $\epsilon$, with the MSE varying at roughly 50\% faster than $\epsilon$. Fig.~\ref{fig:darcy_collage} contains sample input and output fields.

\subsection{2D + 1 Navier Stokes Equation Results}\label{subsec:navier_result}

This 3D vorticity attack was done with no subsampling or grid-search because the model was not trained on subsampled data. Since viewing the 3D data $a$ versus ground-truth was difficult, we only showed the resulting MSE vs $\epsilon$ variations and discussed its implications. Specifically, the MSE is defined 
as $\|\text{FNO}(a+\delta)-\text{PDE-solver}(a+\delta)\|_2^2$, where $\text{FNO}(\cdot)$ denotes the output of the FNO model. Fig.~\ref{fig:pgd_linf_mse_navier} shows the Navier case MSE variation with $\epsilon$, with the MSE varying at roughly 40\% slower than the $\epsilon$. The model showed stronger resistance to input perturbations as the MSE varies less than linearly with the increasing $\epsilon$, possibly due to (i) the use of exact PDE solver as the ground-truth, or (ii) the FNO model is more robust.
However, the variation is more dynamic than the two prior cases.

\section{Conclusion}\label{sec:conclusion}
Our research studies the worst-case sensitivity analysis of FNO models using adversarial examples, based on the rate of changes in MSE with varying perturbation levels.
The rate plots in 1D Burgers and 2D Darcy flow cases were somewhat quadratic while the 2D+1 Navier case seemed cubic. This suggests the potential issue of over-sensitivity; that the model's sensitivity to input perturbations may increase drastically when the underlying problem becomes complex. Our future work includes extending our evaluation tool to other scientific domains and use it to design more robust and generalizable ML-based scientific discovery models. We hope our research findings can inspire future studies on evaluating and improving the adversarial robustness of ML-driven scientific discovery models and inform better design of technically reliable and socially responsible technology.

\section*{Acknowledgement}
This work was supported by the Rensselaer-IBM AI Research Collaboration (\url{http://airc.rpi.edu}), part of the IBM AI Horizons Network (\url{http://ibm.biz/AIHorizons}).

\bibliography{iclr2022_conference,adversarial_learning}
\bibliographystyle{iclr2022_conference}

\newpage

\appendix
\section{Appendix}

\subsection{Setup-Parameters}\label{subsec:params}

\begin{enumerate}
    \item Number of iteration in PGD attack: $K=10$
\item Number of restarts in PGD attack: 10
\item Step size in PGD attack: $ \alpha= \dfrac{\epsilon}{K}$
\item Number of training samples for FNO: 1024
 \item Number of testing samples ($n$): 100
 \item The number of surrogate fields generated ($N$) = 5,000 (Burgers case), 10,000 (Darcy case)
 \item The full grid size $f$  = 1024 
 \item The number of dimensions $d$ = 2 (Burgers case), 3 (Darcy case)
 \item The subsampling rate $s$ = 8
\end{enumerate}

\subsection{Images}

\begin{figure*}[ht]
\centering{\includegraphics[
  width=\linewidth,
  height=0.7\linewidth,
]{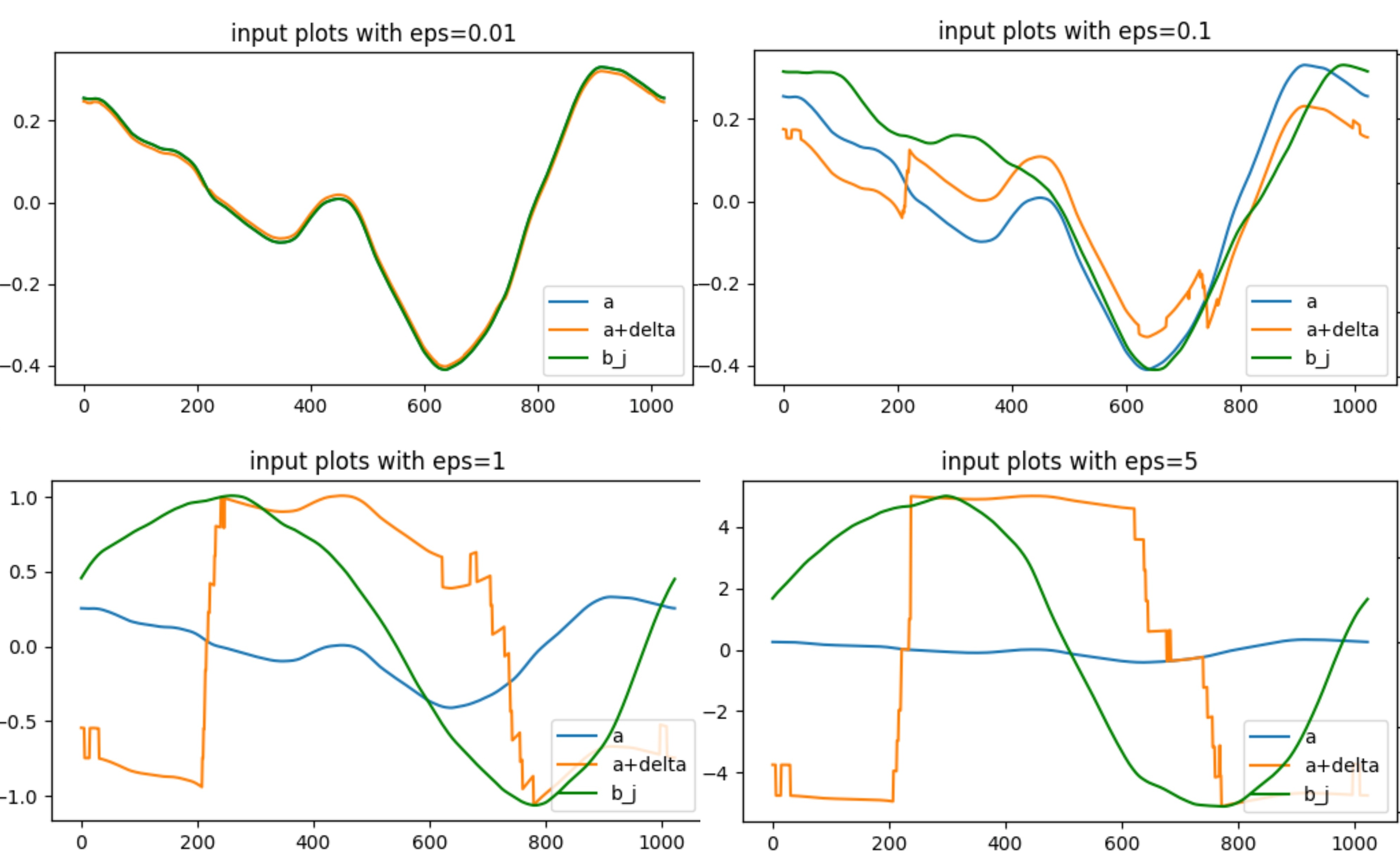}}
\caption{Sample 1D input fields for the Burgers Case with varying $\epsilon$. 
The perturbed sample $a+\delta$ can be very different from its closest surrogate $b_j$.
}\label{fig:burgers_input}
\end{figure*}

\begin{figure*}[ht]
\centering{\includegraphics[
  width=\linewidth,
  height=0.7\linewidth,
]{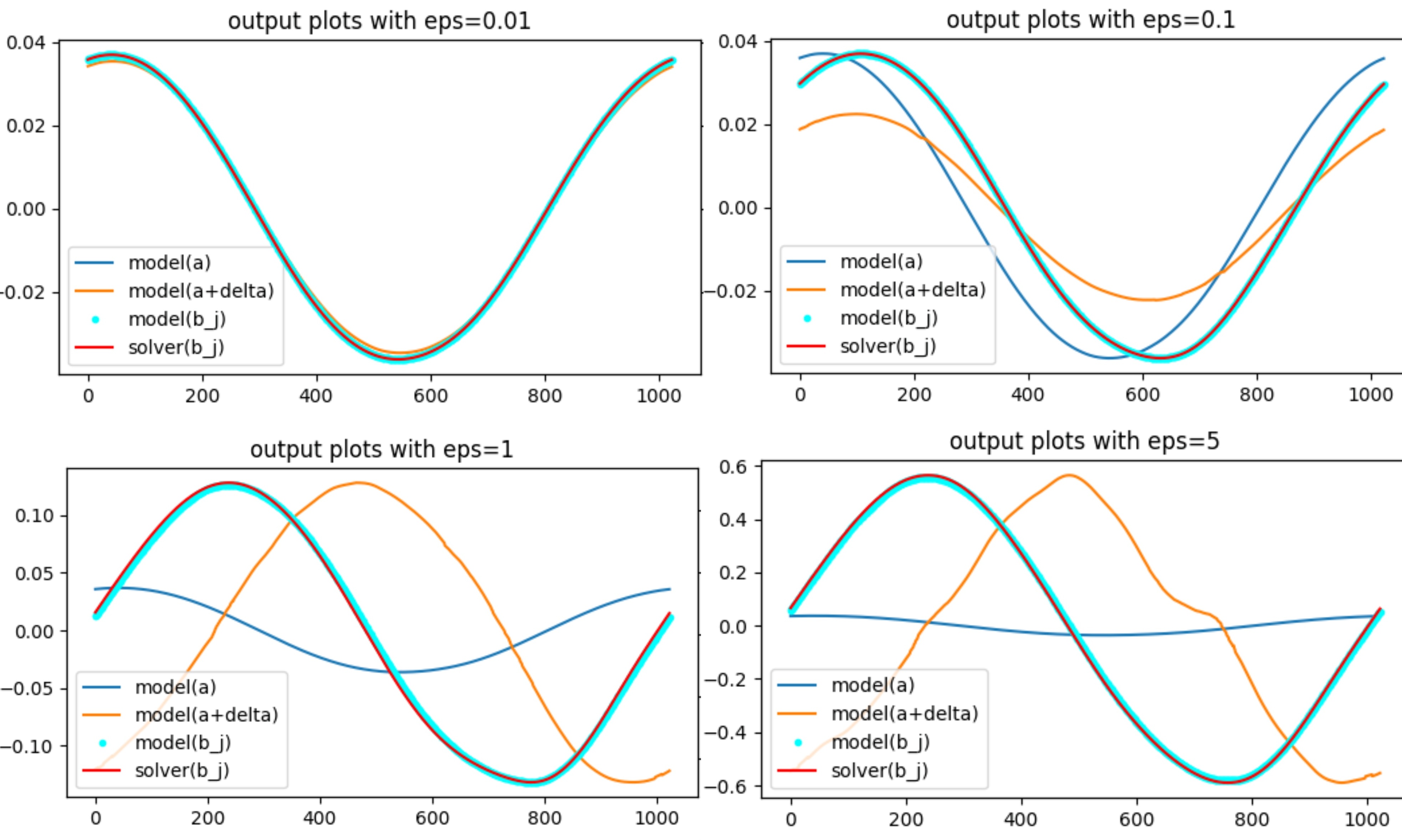}}
\caption{Sample 1D output fields for the Burgers Case with varying $\epsilon$. Notice the FNO output $model(a+\delta)$ could have large difference when compared to the ground-truth $solver(b_j)$ as $\epsilon$ increases. \label{fig:burgers_output}}
\end{figure*}

\begin{figure*}[ht]
\centering{\includegraphics[
  width=\linewidth,
  height=0.7\linewidth,
]{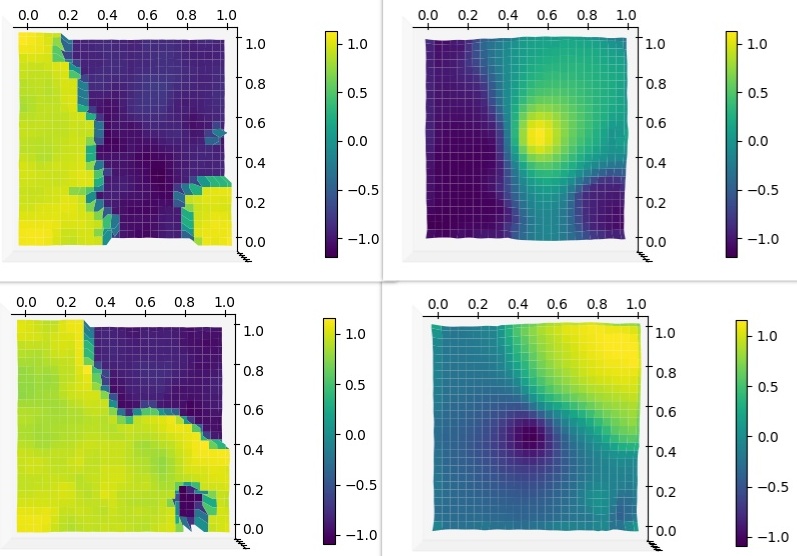}}
\caption{Sample 2D fields for the Navier Case with $\epsilon = 0.01$. \textbf{1st row}: Original input and solution fields. \textbf{2nd row}: closest surrogate field and corresponding solution fields.}\label{fig:darcy_collage}
\end{figure*}

\end{document}

%% file: math_commands.tex
\usepackage{amsmath,amsfonts,bm}

\def\eqref#1{equation~\ref{#1}}

\def\1{\bm{1}}

\DeclareMathAlphabet{\mathsfit}{\encodingdefault}{\sfdefault}{m}{sl}
\SetMathAlphabet{\mathsfit}{bold}{\encodingdefault}{\sfdefault}{bx}{n}

